\makeatletter\def\graphicscache@inhibit{true}\makeatother

\documentclass[letterpaper, 10 pt, conference]{ieeeconf}  %

\IEEEoverridecommandlockouts                              %

\usepackage{amsmath,amssymb,amsfonts}
\usepackage{pifont}
\usepackage{textcomp}
\usepackage{multirow}
\usepackage[hang,flushmargin,symbol]{footmisc}
\usepackage{pgfplots}
\pgfplotsset{compat=newest}
\usepackage{tikz}

\usepackage{algorithm}
\usepackage{algpseudocode}

\usetikzlibrary{fit,arrows,arrows.meta,automata,backgrounds,calc,chains,%
decorations.markings,decorations.pathreplacing,decorations.pathmorphing,%
matrix,positioning,shapes,shapes.geometric,shapes.symbols,spy,trees,tikzmark}

\usepackage{hyperref}
\usepackage{url}
\usepackage{graphicx}
\usepackage{graphicscache} %
\usepackage{booktabs}
\usepackage{acronym}
\usepackage[capitalize]{cleveref}
\usepackage{siunitx}
\usepackage{cite}
\usepackage{flushend}
\usepackage{microtype}

\usepackage{multirow}
\usepackage{gensymb}

\DeclareMathOperator*{\argmin}{arg\,min}

\acrodef{FoV}{Field of View}
\acrodef{LiDAR}{Light Detection and Ranging}
\title{\LARGE \bf
Joint Target-Less Intrinsic and Extrinsic Camera-LiDAR Calibration using Deep Point Correspondences
}

\author{Simon Bultmann, Daniele Cattaneo, and Abhinav Valada%
\thanks{Department of Computer Science, University of Freiburg, Germany}%
\thanks{This work was funded by the German Research Foundation (DFG) Emmy Noether Program under grant 468878300.}%
}

\begin{document}

\maketitle
\thispagestyle{empty}
\pagestyle{empty}

\section{Introduction}
Accurate camera-LiDAR calibration is a fundamental requirement for robotic systems that fuse complementary sensor modalities~\cite{schramm2024bevcar,valada2016convoluted,valada2016towards,bultmann2021real,mohan2026up} to achieve robust perception in real-world environments.
The problem has been studied extensively in the literature using both classical and learning-based methods. Recent learning-based, target-less approaches achieve remarkable performance for extrinsic calibration but assume rectified images with known intrinsics.
In this work, we overcome this limitation and present the first fully target-less pipeline that jointly estimates camera intrinsics (pinhole model with radial-tangential distortion) and camera-LiDAR extrinsics using deep pixel-point correspondences.

Classical calibration methods rely on specially designed calibration targets in controlled setups, such as checkerboards or AprilTags for multi-camera systems~\cite{rehder_kalibr_2016}, hybrid visual-geometric targets~\cite{yan_joint_2023}, or even room-scale environments~\cite{wiesmann_joint_2024}.
While these approaches achieve high accuracy, they are labor-intensive, require expert supervision, and must be repeated whenever sensors move due to vibration, thermal stress, or mechanical wear.
To overcome these limitations, recent research has focused on target-less calibration using naturally occurring scene structure.
Examples include keypoints on objects or persons~\cite{paetzold_camcalib_2022}, geometric primitives such as edges or planes~\cite{li_joint_2023}, and learning-based multi-modal correspondences~\cite{cattaneo2020cmrnetpp,cattaneo_cmrnext_2025,nayak2024ralf}.
Although foundation models for cross-modal matching have emerged~\cite{ren_minima_2025,xue_matcha_2025}, they do not cover sparse LiDAR point clouds due to limited training data.

For camera-LiDAR matching, optical-flow-based methods such as I2DLoc~\cite{yu_i2dloc_2024} and CMRNext~\cite{cattaneo_cmrnext_2025} currently represent the state of the art.
These methods reformulate the pixel-point matching as dense displacement estimation between projected LiDAR scans and RGB images.
Building upon this, MDPCalib~\cite{petek_automatic_2024} introduced a fully automatic target-less extrinsic calibration pipeline by combining motion-based initialization with deep correspondences. However, existing deep correspondence-based camera-LiDAR calibration methods typically assume rectified images with known intrinsics.
In contrast, target-based approaches are able to jointly estimate intrinsic and extrinsic parameters.
In this work, we bridge this gap by extending deep point correspondence-based calibration to jointly recover camera intrinsics and extrinsics without any calibration target or manual initialization.
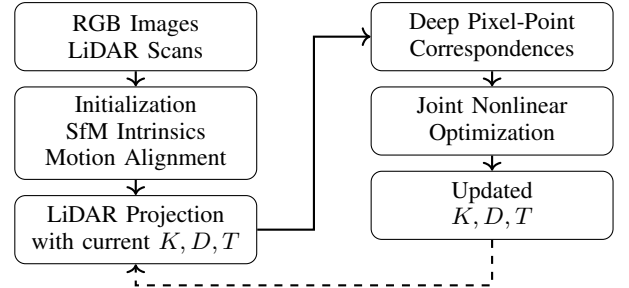
\begin{figure}[t]
\centering
\begin{tikzpicture}[
	xshift=-20cm,
    node distance=0.2cm and 1.5cm,
    block/.style={
        rectangle, draw, rounded corners,
        align=center,
        minimum width=3.2cm,
        minimum height=0.9cm,
        font=\small
    },
    arrow/.style={->, thick},
    dashedarrow/.style={->, thick, dashed}
]

\node[block] (data) {RGB Images\\ LiDAR Scans};
\node[block, below=of data] (init) {Initialization\\ SfM Intrinsics\\ Motion Alignment};
\node[block, below=of init] (proj) {LiDAR Projection\\ with current $K,D,T$};

\node[block, right=of data] (cnn) {Deep Pixel-Point\\ Correspondences};
\node[block, below=of cnn] (opt) {Joint Nonlinear\\ Optimization};
\node[block, below=of opt] (update) {Updated\\ $K,D,T$};

\draw[arrow] (data) -- (init);
\draw[arrow] (init) -- (proj);
\draw[arrow] (proj.east) -- ++(0.75,0) |- (cnn);
\draw[arrow] (cnn) -- (opt);
\draw[arrow] (opt) -- (update);

\draw[dashedarrow] (update.south) -- ++(0,-0.6) -| (proj.south);

\end{tikzpicture}
\vspace{-0.7em}
\caption{Pipeline for joint intrinsic and extrinsic camera-LiDAR calibration. Deep correspondences and nonlinear optimization are iteratively coupled. 
     }
\vspace{-1.5em}
\label{fig:pipeline}
\end{figure}

In summary, we propose a fully target-less pipeline for joint intrinsic and extrinsic camera–LiDAR calibration, including (i) an automatic intrinsic initialization via structure-from-motion, (ii) a generalization of deep camera-LiDAR correspondence learning to raw images with unknown intrinsics, and (iii) a tightly coupled iterative optimization over intrinsics and extrinsics. We further provide a systematic evaluation on the KITTI dataset in an unseen camera setting.

\section{Method}
Figure~\ref{fig:pipeline} illustrates the overall pipeline.
Given synchronized camera images and LiDAR scans, we initialize intrinsics and extrinsics, estimate dense pixel-point correspondences, and iteratively refine all parameters via nonlinear optimization.

\subsection{Camera Model and Projection}

We employ a pinhole camera model with radial-tangential distortion.
Let $\mathbf{X} \in \mathbb{R}^3$ be a LiDAR point transformed into the camera frame by extrinsics
$\mathbf{T} = \bigl(\begin{smallmatrix}
\mathbf{R} & \mathbf{t}\\
\mathbf{0}^\top & 1
\end{smallmatrix}\bigr) \in \mathbb{SE}(3)$, with $\mathbf{R}\in\mathbb{SO}(3)$, corresponding quaternion $\mathbf{q_R}$, and $\mathbf{t}\in\mathbb{R}^3$:
\begin{equation}
\mathbf{X}_c = \mathbf{R}\mathbf{X} + \mathbf{t}\,.
\label{eq:proj_start}
\end{equation}
Then, the normalized image coordinates are given as
\begin{equation}
x = X_c / Z_c,\quad y = Y_c / Z_c\,.
\end{equation}
Using the distortion parameters $\mathbf{D}=[k_1,k_2,t_1,t_2,k_3]^\top$,
radial distortion is modeled with $r^2 = x^2 + y^2$ as
\begin{equation}
f(r) = 1 + k_1 r^2 + k_2 r^4 + k_3 r^6\,.
\end{equation}
Together with tangential distortion, distorted coordinates are
\begin{align}
x_d &= x f(r) + 2 t_1 xy + t_2 (r^2 + 2x^2)\,, \\
y_d &= y f(r) + 2 t_2 xy + t_1 (r^2 + 2y^2)\,.
\end{align}
Finally, pixel coordinates are obtained via
\begin{equation}
\mathbf{u}
=
\mathbf{K}
\begin{bmatrix}
x_d \\
y_d \\
1
\end{bmatrix},
\text{where}\;
\mathbf{K} =
\begin{bmatrix}
f_x & 0   & c_x \\
0   & f_y & c_y \\
0   & 0   & 1
\end{bmatrix}\,,
\label{eq:proj_end}
\end{equation}
with focal lengths $f_x, f_y$ and optical center $\mathbf{c} = [c_x, c_y]^\top$.

\subsection{Joint Nonlinear Optimization}

Given a set of correspondences $\mathcal{C}=\{(\mathbf{X}_i,\mathbf{u}_i)\}$, we jointly optimize intrinsics $\boldsymbol{\theta}_K = (\mathbf{K},\mathbf{D})$ and extrinsics $\boldsymbol{\theta}_T = (\mathbf{q_R},\mathbf{t})$ by minimizing reprojection error:
\begin{equation}
\boldsymbol{\theta}^* = \argmin_{\boldsymbol{\theta}_K,\boldsymbol{\theta}_T}
\sum_i \rho\left(\left\|
\mathbf{u}_i -
\Pi(\mathbf{X}_i;\boldsymbol{\theta}_K,\boldsymbol{\theta}_T)
\right\|^2\right)\,,
\end{equation}
where $\Pi(\cdot)$ denotes the full projection model Eqs. \eqref{eq:proj_start}--\eqref{eq:proj_end} and $\rho(\cdot)$ is the robust Cauchy loss~\cite{petek_automatic_2024}.
The optimization is performed using nonlinear least squares~\cite{Agarwal_Ceres_Solver_2022}.
We further add a weak prior on the principal point $\mathbf{c}$, anchored at the SfM initialization $\mathbf{c}_0$, that helps to stabilize the optimization without fixing the principal point.

\subsection{Initialization}

To initialize intrinsics without calibration targets, we run structure-from-motion on a small subset $N_{\text{init}}$ of images with sufficient motion using COLMAP~\cite{schoenberger2016sfm}, estimating focal length, principal point, and radial-tangential distortion.
Extrinsics are initialized using the motion-alignment strategy of MDPCalib~\cite{petek_automatic_2024}, aligning visual and LiDAR odometry.

\subsection{Learning Pixel-Point Correspondences with Unknown Intrinsics}

To generalize deep camera-LiDAR correspondence learning to raw images, we extend the CMRNext training pipeline~\cite{cattaneo_cmrnext_2025} to operate under unknown camera intrinsics and distortion.
In contrast to the original pipeline, which adds noise only to the extrinsics, we additionally vary the intrinsic parameters when projecting LiDAR points during training.
The parameters are perturbed via uniform random noise, and multiple network instances are trained with different noise magnitudes following~\cite{cattaneo_cmrnext_2025}. To ensure physically plausible camera models, distortion parameters are sampled under sign- and scale-consistency constraints.

To augment training data with diverse camera distortion models, we additionally generate synthetically distorted images by remapping between cameras.
We define a dense mapping between two camera models
$\mathcal{C}_1=(\mathbf{K}_1,\mathbf{D}_1)$ and $\mathcal{C}_2=(\mathbf{K}_2,\mathbf{D}_2)$.
For each pixel $\mathbf{u}_2$ in the target camera, we compute the corresponding pixel in the source camera as
\begin{equation}
\mathbf{u}_1
=
\Pi\!\left(
\Pi^{-1}(\mathbf{u}_2;\mathbf{K}_2,\mathbf{D}_2);
\mathbf{K}_1,\mathbf{D}_1
\right),
\end{equation}
where $\Pi(\cdot)$ and $\Pi^{-1}(\cdot)$ denote the forward projection and unprojection operators,
respectively.
Pixels are considered valid only if the corresponding normalized ray lies within the maximum
valid distortion radius $r_{\max}$ of both camera models~\cite{leotta_maximum_2022}.
The resulting dense remapping is applied using standard image warping.

\subsection{Coupled Iterative Refinement}

Unlike prior work, which relies on PnP to estimate extrinsics from correspondences under fixed intrinsics, we tightly couple correspondence estimation and nonlinear optimization.
At each iteration, correspondences predicted by CMRNext, using models trained with decreasing initial noise scales, are fed into the joint optimization to update intrinsics and extrinsics.
After $N_{\text{intr}}=5$ iterations, the intrinsic parameters are fixed and the optimization continues until $N_{\text{total}}=11$ iterations, further refining the extrinsics. This strategy stabilizes convergence and improves robustness. The stop iterations are found using a hyperparameter search.

\section{Experimental Evaluation}
\begin{table}[t]
\centering
\caption{Joint intrinsic and extrinsic calibration results on KITTI (right camera, unseen setup).
Mean $\pm$ std over 100 runs.}\vspace{-1em}
\label{tab:results}
\setlength{\tabcolsep}{3pt}
\begin{tabular}{lccc}
\toprule
Method & Transl. [cm] & Rot. [$^\circ$] & Reproj. [px] \\
\midrule
MDPCalib~\cite{petek_automatic_2024} 
      & 2.94 & 0.14 & -- \\
\midrule
Ours (full, GT init)      
      & $\textbf{2.27} \pm \textbf{0.09}$ & $\textbf{0.106} \pm \textbf{0.019}$ & $\textbf{1.14} \pm \textbf{0.24}$ \\
Ours (full, COLMAP init) 
      & $2.34 \pm 0.12$ & $0.108 \pm 0.028$ & $1.25 \pm 0.38$ \\
Ours (no ppp, GT init) 
      & $2.48 \pm 0.24$ & $0.241 \pm 0.117$ & $4.11 \pm 1.84$ \\
\bottomrule\\[-.9em]
\end{tabular}
\footnotesize
\raggedright
init: adding $\pm 10\%$ uniform random noise to resp. initial intr. params.\\
no ppp: without weak prior on principal point.
\vspace{-.3cm}
\end{table}

Our evaluation follows the protocol of~\cite{petek_automatic_2024,cattaneo_cmrnext_2025}, where the rectified left camera of KITTI~\cite{kitti_geiger_2012} sequences 01--21, together with one (rectified) camera of Argoverse~\cite{argoverse_chang_2019} and Pandaset~\cite{pandaset_xiao_2021}, respectively, are used for training.
We further add raw KITTI images, where available, and our synthetic distortion augmentation.
Evaluation is performed on the raw right camera of KITTI sequence~00, representing a completely unseen camera-LiDAR pair on an unseen sequence.

Extrinsic accuracy is measured via translation error and shortest-arc rotation error.
Intrinsic accuracy is evaluated using reprojection error: given normalized rays sampled uniformly over the image, we compare pixel projections under ground-truth and estimated parameters, reporting the median reprojection error over the sampled rays. This metric is more meaningful than comparing intrinsic parameters directly, as they have heterogeneous numerical scales and are partially coupled, whereas reprojection errors directly measure the geometric accuracy.
All results are averaged over 100 runs, with mean and standard deviation reported.

Our results in Table~\ref{tab:results} show that joint intrinsic-extrinsic calibration achieves lower extrinsic error than prior methods while additionally recovering accurate intrinsics.
To highlight robustness to different initializations of intrinsic parameters, we add $\pm 10\%$ uniform random noise to both the ground-truth (GT) and the SfM initialization (COLMAP).
Ablations further demonstrate the benefit of a weak prior on the principal point: since it is only weakly observable from reprojection error and coupled with camera rotation, unconstrained principal point updates tend to be compensated by rotations, leading to increased rotation and intrinsic reprojection error. 

\section{Conclusion}
We presented the first target-less approach for joint intrinsic and extrinsic camera-LiDAR calibration using deep point correspondences.
By extending correspondence learning to raw images and tightly coupling it with nonlinear optimization, our method removes the need for pre-calibrated intrinsics without sacrificing extrinsic accuracy.
Future work includes evaluation on real-world robotic platforms as well as automatic miscalibration detection and online recalibration.

\addtolength{\textheight}{-10cm}   %
\bibliographystyle{IEEEtranDOI}
\bibliography{references.bib}

\end{document}